\documentclass[utf8]{arxivmanuscript}
\usepackage{url}
\usepackage{hyperref}
\hypersetup{hidelinks}
\AtBeginDocument{\def\doi#1{doi:\href{https://doi.org/#1}{\nolinkurl{#1}}}}
\usepackage{lineno}
\usepackage{microtype}
\usepackage{subcaption}
\usepackage[onehalfspacing]{setspace}
\usepackage{graphicx}
\graphicspath{{./}}
\usepackage{newunicodechar}
\usepackage{xcolor}
\usepackage{float}
\newunicodechar{≥}{\ensuremath{\geq}}
\newunicodechar{×}{\ensuremath{\times}}
\newunicodechar{→}{\ensuremath{\rightarrow}}
\usepackage{fvextra}
\usepackage{longtable}
\usepackage{booktabs}
\usepackage{array}
\def\keyFont{\fontsize{8}{11}\helveticabold}
\def\firstAuthorLast{Mobini {et~al.}}
\def\Authors{
Melika Mobini\,$^{1,*}$,
Andres Algaba\,$^{1}$,
Vincent Ginis\,$^{1,2}$
}
\def\Address{
$^{1}$Vrije Universiteit Brussel, Brussels, Belgium\\
$^{2}$Harvard University (SEAS), Cambridge, MA, United States
}
\def\corrAuthor{Melika Mobini}
\def\corrEmail{Melika.Mobini@vub.be}

\begin{document}
\onecolumn
\raggedbottom % avoid stretched vertical gaps when [H] figures move to the next page
\renewcommand{\floatpagefraction}{0.6} % let large [p] figures (e.g. Fig. 3) take their own page
\firstpage{1}
\title[Reasoning Effort and Reference Verifiability]{Higher Reasoning Effort Improves Reference Verifiability and Changes Reasoning Summaries in GPT-5.5}

\author[\firstAuthorLast]{\Authors}
\address{}
\correspondance{}

\extraAuth{}

\maketitle

\begin{abstract}

Large language models are increasingly being used in different stages of scientific knowledge production, including the suggestion of references. These models can operate at different levels of reasoning effort and provide summaries of their reasoning processes. This raises two questions: Does increasing reasoning effort improve the verifiability of generated references, and do the resulting reasoning summaries reveal how the model's behavior changes with greater effort? We examined summaries from GPT-5.5 as it generated scholarly references for 1,000 papers under paired low- and high-effort settings. Generated references were matched to external bibliographic records, and reasoning summaries were scored for four predefined features and explored for additional features discovered by GPT-5.5. High effort increased the mean paper-level resolver-verified reference rate from 78.3\% to 84.8\%, and produced longer reasoning summaries. All four predefined features received higher scores under high effort. The discovered features changed less uniformly: some appeared equally often in both conditions, whereas others were more common under high effort. Within each effort condition, neither feature set showed a consistent increase in verification rates with higher scores.

\tiny
\keyFont{\section{Keywords:} large language models, reasoning effort, reasoning summaries, reference generation, citation hallucination, bibliographic verification}
\end{abstract}

%%%%%%%%%%%%%%%%%%%%%%%%%%%%%%%%%%%%%%%%%%%%%%%%%%%%%%%%%%%%%%%%%%%%

\section{Introduction}

Large language models (LLMs) can recommend scholarly references, but these references may point to nonexistent publications, contain incorrect bibliographic details, or be unrelated to the requested topic \citep{walters2023,byun2024}. Such errors can make it harder for readers to locate relevant studies and verify the evidence associated with recommended references. Reference generation therefore provides a useful setting for examining model output quality alongside the reasoning described during generation. However, reference quality is not a single dimension. \citet{walters2023} distinguish between references to nonexistent publications and bibliographic errors in references to existing works. A generated reference may correspond to a real publication while containing incorrect author names, titles, publication years, or other bibliographic details. \citet{byun2024} suggested that improvements in title and author accuracy do not necessarily make reference recommendations more relevant. Similarly, \citet{lehr2024chatgpt} assessed correctness, completeness, and topical relevance separately. These studies showed why reference existence, bibliographic accuracy, and relevance should be evaluated separately. Identifying a publication and assessing what it supports are also different tasks. CiteME asked models to identify the paper referred to in a scientific text excerpt \citep{press2024citeme}. In medical question answering, \citet{wu2025medicalreferences} distinguished valid source URLs from sources that support individual statements or the response as a whole. Researchers have also proposed methods for detecting nonexistent or incorrectly specified references. \citet{agrawal2024} asked models direct and indirect questions about generated references to check their consistency, showing that some hallucinated references can be identified without external sources. CiteCheck, in contrast, retrieved candidate publications and checked whether their bibliographic details matched the generated citation \citep{khajavi2026}. These approaches illustrate different ways of assessing the reliability of generated references. Related research has examined broader patterns in LLM-generated reference lists, including differences across scientific fields and overlap with observed bibliographies \citep{algaba2025,mobini2026}. In a large-scale study, \citet{algaba2025} found that reference existence rates varied across scientific fields and that generated lists had limited overlap with the focal papers' bibliographies. \citet{mobini2026} compared citation networks built from generated and ground-truth references. They found that the networks had similar structural properties but could be distinguished more accurately using embeddings of the cited publications' titles. Reference existence, bibliography overlap, network structure, and content similarity therefore capture different aspects of generated reference lists.

Increasing the computation allocated to a model during inference can improve its performance on reasoning tasks \citep{openai2024reasoning}. However, the benefit depends on the difficulty of the task and how the additional computation is used \citep{snell2025scaling}. In scholarly reference generation, this raises the question of whether greater reasoning effort improves the reliability of generated references. Reasoning models also provide summaries of their reasoning processes, which may help us understand how their approach changes with reasoning effort. However, these summaries do not necessarily reflect how models actually produce their answers. Similar concerns have been raised about chain-of-thought explanations. For example, \citet{turpin2023} showed that such explanations could omit information that influenced model predictions. Or \citet{lanham2023} found that the extent to which final answers depended on the stated reasoning varied across models and tasks. Similarly, \citet{chen2025} found that Claude 3.7 Sonnet and DeepSeek R1 often did not mention hints that influenced their answers. Although these studies examined chain-of-thought reasoning rather than the reasoning summaries considered here, they highlight the limitations of interpreting visible reasoning as a complete account of model behavior. Nevertheless, such text may still provide useful information about model behavior and performance. \citet{guan2026monitoring} examined whether visible reasoning could help assess model behavior without assuming that it reveals all internal computation. Across three benchmarks and five open-weight Qwen3 and gpt-oss models, \citet{algaba2026observability} compared assessments of answer correctness based on model responses, model-written summaries of reasoning traces, full reasoning traces, and internal signals. They found that, when the original prompt was unavailable, summaries provided most of the information available in full traces for ranking answer correctness, whereas summaries added little when the prompt was available. More generally, LLM-based coding can show substantial agreement with human judgments, although this agreement varies across models and evaluated features \citep{dunivin2024}. These findings motivate examining reasoning summaries as observable records of model behavior. 

The behaviors expressed in these summaries may help us understand how a model approaches a task, but what they reveal does not necessarily tell us whether its answers are correct. Expressions of uncertainty are one example. \citet{kalaiEtAl2025} argued that common training and evaluation methods can encourage models to guess instead of admitting uncertainty, which may lead to confident but incorrect answers. \citet{kadavath2022know} examined whether models could estimate when their answers were correct and found limitations in the accuracy of these estimates across different data distributions. However, a model's ability to estimate its own correctness is different from how it expresses uncertainty in reasoning text. \citet{vanhoyweghen2025lexical} found that uncertainty-related wording in chain-of-thought traces was associated with incorrect answers. We therefore examine whether epistemic caution expressed in reasoning summaries is associated with reference verification, without assuming that more cautious language means more accurate references.

\begin{figure}[t]
    \centering
    \includegraphics[width=\textwidth]{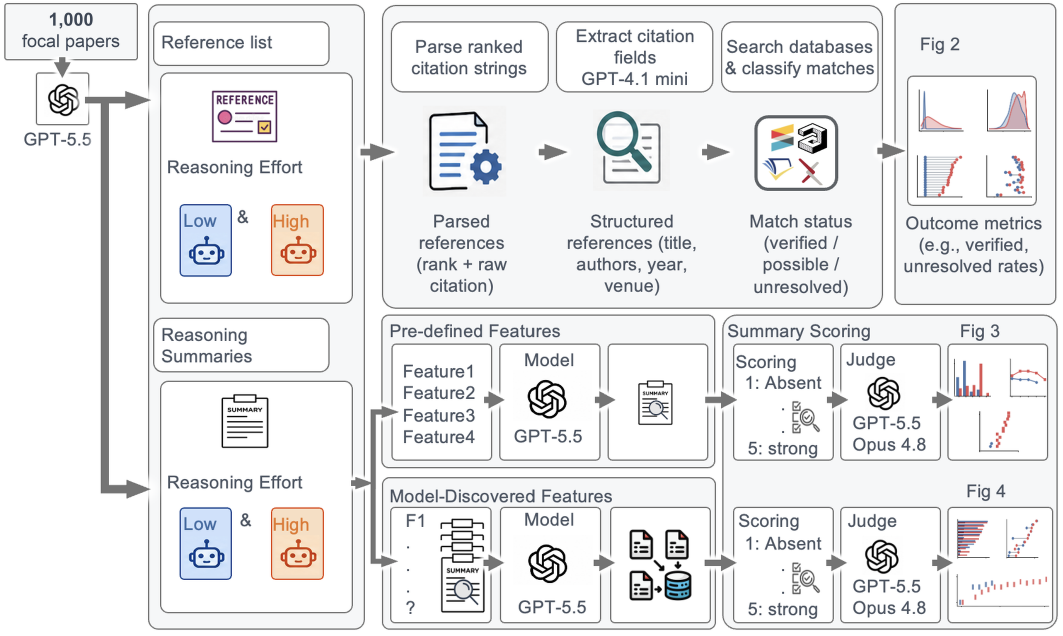}
     \caption{\textbf{Study design and analysis pipeline.}
    GPT-5.5 generates references and a reasoning summary for each of 1,000 focal papers under low and high reasoning effort.
    \textit{Top:} Generated references are parsed, their metadata extracted using GPT-4.1 mini, and they are matched to external bibliographic records to determine verification status and metadata agreement. The resulting paper-level verified-reference rates are compared between low- and high-effort settings, overall and by research field.
    \textit{Bottom:} Reasoning summaries are scored on predefined features and on model-discovered features derived by GPT-5.5, using a 1–5 scale from absent to strong, with GPT-5.5 and Claude Opus 4.8 serving as judges. (Figures ~\ref{fig3}--\ref{fig4}).}
    \label{fig:pipeline}
\end{figure}

Previous studies have examined the reliability of LLM-generated references and the behaviors expressed in model reasoning, but how reasoning effort affects reference verifiability and whether observable reasoning behaviors are associated with verification outcomes remain less understood. We address these questions through a paired experiment in which GPT-5.5 generated 20 scholarly references for each of 1,000 focal papers under low and high reasoning effort. We compare reference-verification outcomes across the two conditions and examine the accompanying reasoning summaries using four predefined features and additional features identified by the model. We assess how these features differ across effort levels, how their expression varies with summary length, and whether they are associated with reference verification. Figure~\ref{fig:pipeline} summarizes the study design.

%%%%%%%%%%%%%%%%%%%%%%%%%%%%%%%%%%%%%%%%%%%%%%%%%%%%%%%%%%%%%%%%%%%%

\section{Materials and Methods}

\subsection{Citation generation and reference verification}

We selected 1,000 focal papers from the open dataset of \citet{algaba2025}, sampled from SciSciNet \citep{lin2023sciscinet}. Eligible papers were those published in first-quartile journals between 1999 and 2021 that had 3--54 references, at least one received citation, a defined top-level field, a valid DOI, and an available abstract. For each focal paper, we asked GPT-5.5 to generate 20 relevant scholarly references under low and high reasoning effort. The prompt (Supplementary Section~\ref{app:prompt-generation}) included the paper's title, authors, publication year, venue, and abstract. The prompt did not require references to exist, predate the focal paper, or appear in its bibliography. No citation style or structured output format was specified, allowing us to assess reference generation without explicitly imposing these constraints.
The focal paper's reference list was excluded from the prompt, and the model had no access to external retrieval tools.
We retained one response per paper and condition, holding the model, prompt, requested reference count, and evaluation procedure constant. Supplementary Table~\ref{tab:config} gives model identifiers and request settings.
GPT-4.1 mini extracted bibliographic metadata from each generated reference (Supplementary Section~\ref{app:prompt-enrichment}). We retrieved candidate records sequentially from OpenAlex \citep{priem2022openalex}, Crossref \citep{hendricks2020crossref}, Semantic Scholar \citep{kinney2023semanticscholar}, and arXiv.
After searching each source, we compared the generated references with the retrieved candidate records using a weighted score based on title and author similarity, calculated with RapidFuzz \citep{bachmann2026rapidfuzz}.
A generated reference was classified as a \texttt{verified-paper} if at least one of these candidates had title similarity of at least \(0.90\), author similarity of at least \(0.50\), and an exact match between the normalized first-author surnames. If no candidate met these criteria, the reference was classified as a \texttt{possible-paper} when at least one of the three candidates had title similarity greater than \(0.80\) and author similarity greater than \(0.40\). When candidates were retrieved but none met either set of thresholds, the reference was classified as \texttt{unresolved}. Failure to identify an acceptable match does not establish that a publication does not exist.

%%%%%%%%%%%%%%%%%%%%%%%%%%%%%%%%%%%%%%%%%%%%%%%%%%%%%%%%%%%%%%%%%%%%

\subsection{Reasoning summaries and predefined features}

The model provided reasoning summaries do not capture its full internal reasoning. A summary was unavailable for 138 low-effort observations and 101 high-effort observations. References were nevertheless generated for these observations. We defined four features: epistemic caution, temporal-constraint awareness, reference-verifiability checking, and metadata verification. Epistemic caution captures explicit recognition of uncertainty and avoidance of assuming accuracy. Temporal-constraint awareness captures explicit checking that a proposed reference does not postdate the focal paper. Reference-verifiability checking captures consideration of whether a proposed publication exists or can be verified. Metadata verification captures checks of specific bibliographic attributes, including authors, titles, or other bibliographic metadata. We asked GPT-5.5 to score the extent to which each feature was expressed in the reasoning summaries. The features were scored separately because they capture different behaviors. The judge assessed only what was expressed in each summary, without external knowledge. Each feature received an integer score on a five-point ordinal scale: 1 indicates no observable expression of the behavior; 2, a weak or generic expression; 3, implicit expression or indirect adaptation of the generation strategy; 4, explicit checking or self-questioning; and 5, sustained or repeated checking. Supplementary Table~\ref{tab:feature_score_examples} provides examples illustrating the assigned feature scores. Scores of at least 3 were treated as feature presence. The primary judge was GPT-5.5 (settings in Supplementary Table~\ref{tab:config}). To assess whether the results depended on this judge, all summaries were also scored by Claude Opus 4.8 using the same feature definitions and rubric (Supplementary Section~\ref{app:prompt-predefined}); the comparison results are reported in Supplementary Section~\ref{app:judge}. Observations without a summary were retained and assigned a score of 1 for all four features. This assignment indicates that the behavior is not observable in the saved record; it does not establish its absence from the model's internal reasoning.

%%%%%%%%%%%%%%%%%%%%%%%%%%%%%%%%%%%%%%%%%%%%%%%%%%%%%%%%%%%%%%%%%%%%

\subsection{Model-discovered feature analysis}

We complemented the predefined-feature analysis with a separate exploratory stage in which GPT-5.5 identified additional features expressed in the reasoning summaries. Feature discovery was conducted separately for the low- and high-effort conditions. Within each condition, the 1,000 focal-paper records were divided into five batches of 200 papers each. This batch size matched the 200-paper batches used for reference generation, while keeping the combined summaries within the model's context-window limits. GPT-5.5 received the summaries from each batch and was instructed to identify reasoning features, providing a concise name and operational definition for each (Supplementary Table~\ref{tab:config}). In a separate GPT-5.5 call (Supplementary Section~\ref{app:prompt-model-scoring}), each paper was scored only against features identified in its own condition and batch. Thus, each feature was evaluated across 200 papers, and papers outside that batch were not considered as lacking it. For each paper, the judge received its reasoning summary and the feature definitions for its batch. It assessed only behaviors expressed in the summary, without external knowledge. Each feature was scored using the same five-point scale as the predefined features, with empty summaries assigned a score of 1 for all features. 
After scoring, we matched features with similar meanings across effort conditions. We retained only features with a matching low--high pair in all five batches. Each paper's score came from its corresponding batch-specific feature, with scores of 3 or higher indicating presence. Supplementary Table~\ref{tab:harmonized_model_features} lists these features and gives representative definitions produced by GPT-5.5.

%%%%%%%%%%%%%%%%%%%%%%%%%%%%%%%%%%%%%%%%%%%%%%%%%%%%%%%%%%%%%%%%%%%%

\section{Results}

\subsection{Summary length and reference verification}

Across the 1,000 focal papers, GPT-5.5 generated 39,986 of the 40,000 requested references (20,000 low effort; 19,986 high effort). The 14 missing references resulted from one high-effort response that returned six instead of 20 references. Under low effort, 15,651 references (78.3\%) were verified, 9.35\% were possible matches, and 12.40\% were unresolved. Under high effort, 16,946 references (84.8\%) were verified, 8.45\% were possible matches, and 6.76\% were unresolved. These are pooled descriptive differences across generated references. Across 1,000 focal papers, the mean paired increase in verification rate was 6.53 percentage points (95\% CI 5.78--7.28) \citep{efron1993bootstrap}. Retrieval and matching details together with metadata classification results and source-level retrieval counts appear in Supplementary Sections~\ref{app:reference-matching}, ~\ref{app:metadata-agreement-results} and~\ref{app:resolver-source-results}.
Supplementary Figures~\ref{fig:temporal_patterns} and~\ref{fig:resolver_scores} also show the publication-year patterns behind the temporal outcome and the candidate-score boundaries behind the high-effort resolver classifications, respectively.

\begin{figure}[p]
    \centering
    \includegraphics[width=0.96\textwidth]{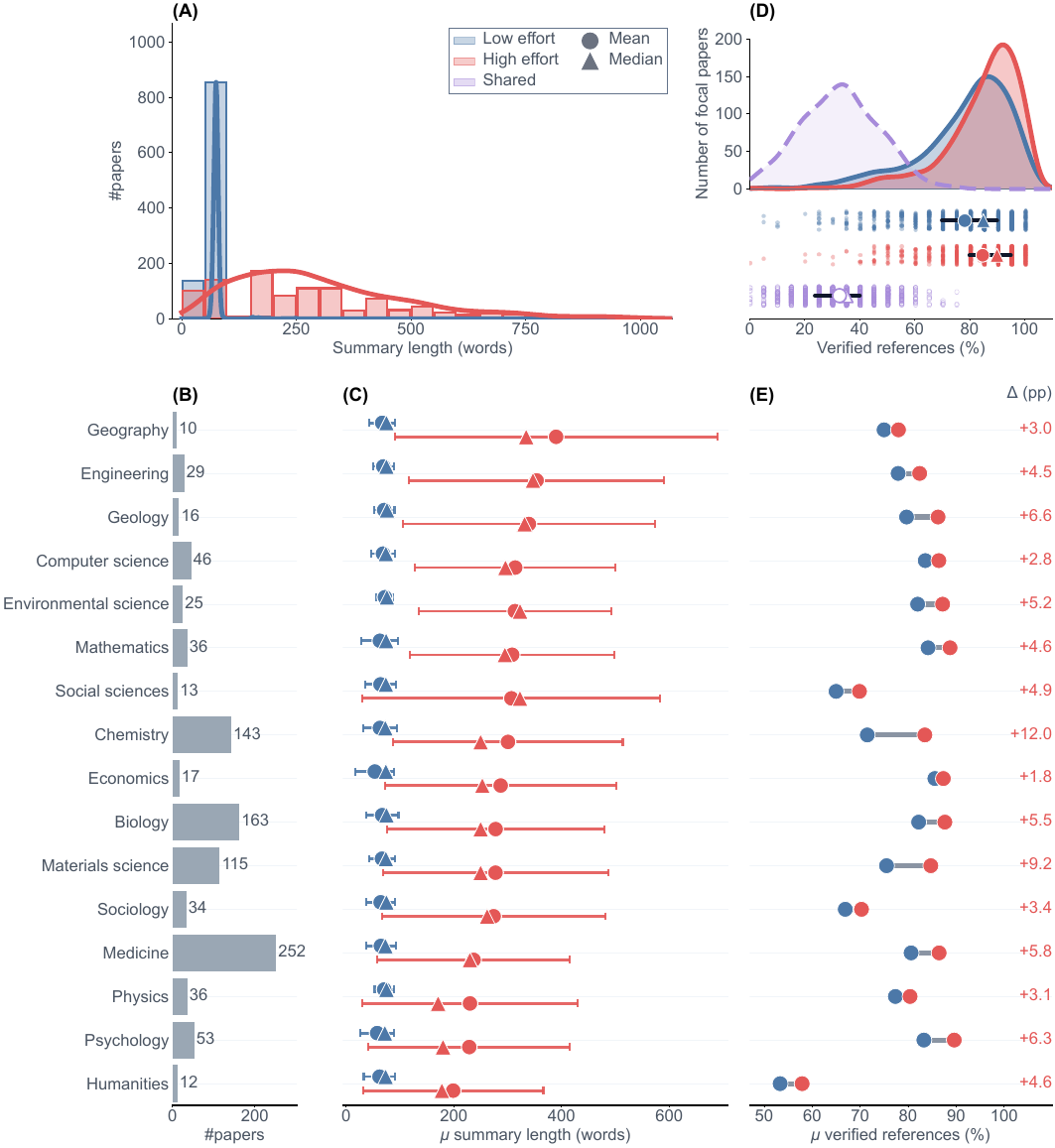}
    \caption{\textbf{Effect of reasoning effort on summary length and verified-reference rates.} Paired observations for 1,000 focal papers. \textbf{(A)}, Distribution of reasoning-summary length in words. Curves are smoothed density estimates of summary lengths, each scaled to the tallest histogram bar of its condition. \textbf{(B)}, Number of focal papers in each research field; these are the denominators for panels (C) and (E). \textbf{(C)}, Summary length by field. Horizontal bars show the mean $\pm$1 between-paper standard deviation. \textbf{(D)}, Distribution of paper-level verified-reference rates under low and high reasoning effort. The purple distribution shows, for each paper, the percentage of its low-effort references that also appear in the high-effort list and are verified. Upper curves are count-scaled density estimates (dashed for shared). Each point below represents one focal paper, with hollow markers for shared rates. Thick black bars show the IQR. \textbf{(E)}, Mean verified-reference rate by field. $\Delta$ (pp) gives the high-minus-low difference in percentage points.}
\label{fig:effort_inflation}
\end{figure}

Higher reasoning effort increased both summary length and the proportion of references verified by the resolver, with variation across scientific fields. Across all 1,000 focal papers, median summary length increased from 74 words under low effort to 248 words under high effort (Figure~\ref{fig:effort_inflation}A). Runs without a summary were counted as having zero words. Figure~\ref{fig:effort_inflation}B shows the number of focal papers per field. Fields with fewer than ten papers were grouped into Humanities (Art, History, and Philosophy; \(n=12\)) and Social sciences (Business and Political science; \(n=13\)). In Figure~\ref{fig:effort_inflation}C we illustrate the mean summary length per field. Mean lengths ranged from 53 to 72 words under low effort and under high effort mean length ranged from 199 words in Humanities to 390 words in Geography. Means and medians were similar in some fields, including Engineering and Medicine, whereas under high effort medians exceeded means in fields such as Social Sciences and Environmental Science.
These differences describe the length of the saved text and do not measure how much internal reasoning a field requires.
The mean paper-level verified-reference rate increased from 78.3\% under low effort to 84.8\% under high effort (Figure~\ref{fig:effort_inflation}D). The median increased from 85\% to 90\%, and the inter-quartile range (IQR) shifted from 70--90\% to 80--95\%. For each focal paper, we also identified references that appeared in both the low- and the high-effort list (7,334 pairs). The purple distribution in Figure~\ref{fig:effort_inflation}D shows the percentage of a paper's low-effort references that are shared and verified. On average this was 32.6\%, so roughly a third of each low-effort list was both repeated under high effort and verified. Figure~\ref{fig:effort_inflation}E shows the mean verified-reference rate per field. Low-effort means ranged from 53.3\% to 85.6\%. High effort increased the mean verified-reference rate in every displayed field, with gains ranging from 1.8 to 12.0 percentage points.

%%%%%%%%%%%%%%%%%%%%%%%%%%%%%%%%%%%%%%%%%%%%%%%%%%%%%%%%%%%%%%%%%%%%

\subsection{Predefined-feature scores, citation outcomes, and summary length}

\begin{figure}[p]
    \centering
    \includegraphics[width=\textwidth]{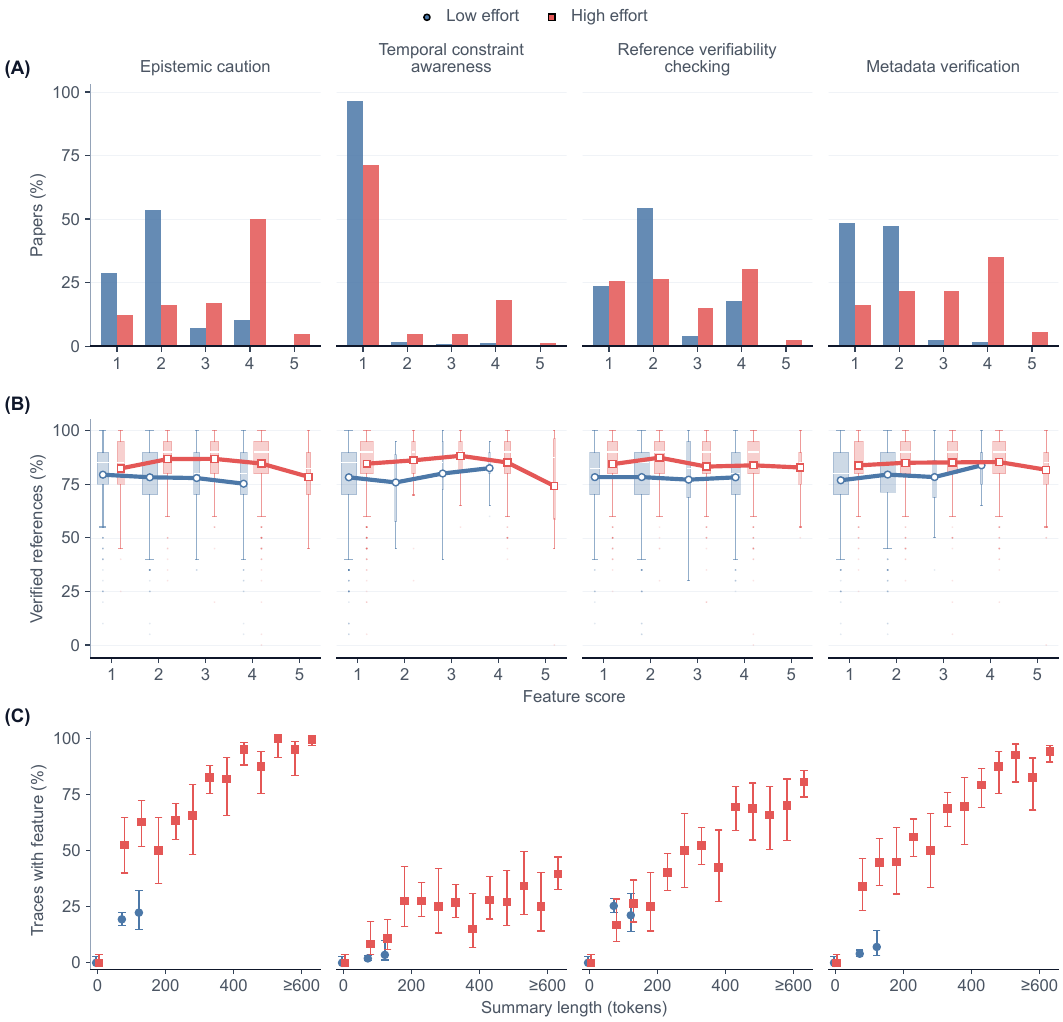}
    \caption{\textbf{Predefined feature expression in reasoning summaries.} Columns show four  predefined features, each scored 1--5 by a GPT-5.5 judge. A feature counts as present when its score is 3 or higher.
    \textbf{(A)}, Percentage of papers at each score, comparing effort conditions.
    \textbf{(B)}, Paper-level verified-reference rates grouped by feature score within each condition. Boxes show medians and IQRs; lines connect group means.
    \textbf{(C)}, Percentage of paper-condition observations in which the feature is present, in 50-token summary-length bins. Error bars show Wilson 95\% confidence intervals \citep{wilson1927probable}. Missing summaries are plotted at 0 tokens; their zero feature prevalence follows from this coding. Lengths above 600 tokens are pooled, and bins with fewer than five observations are omitted.}
    \label{fig3}
\end{figure}

Figure~\ref{fig3} summarizes feature-score distributions, their relationships with reference verification, and their variation with summary length. Scores for all four features were higher under high reasoning effort (Figure~\ref{fig3}A). The largest shifts were observed for epistemic caution and metadata verification. Under low effort, 82.5\% of observations received scores of 1 or 2 for epistemic caution, and 95.9\% received these scores for metadata verification. Temporal-constraint awareness was rarely expressed under low effort: 96.3\% of observations received a score of 1. However, high-effort scores were distributed across a wider range of the scale. Scores of 5 occurred only under high effort and were uncommon. Reference-verifiability checking also shifted toward higher scores under high effort, although less sharply than epistemic caution and metadata verification. Claude Opus 4.8 also scored the same feature expressions (Supplementary Figure~\ref{fig:judge_robustness}A). This judge assigned lower scores overall, but all four features again scored higher under high effort.
High effort had a higher verified proportion at every score level observed in both conditions. But within each effort condition, verification rates showed no consistent increase with feature score (Figure~\ref{fig3}B). The mean curves are broadly flat with small movements, even though high-effort rates remain above low-effort rates at shared score levels. To examine whether an association appeared only at the extremes of the outcome distribution, we compared papers at or below the 10th percentile of verified-reference rate with those at or above the 90th percentile within each condition (Supplementary Table~\ref{tab:decile}). 
The plotted mean verification rates show no consistent increase across predefined-feature score levels within either effort condition. These descriptive comparisons do not establish whether these behaviors contributed to the difference in citation outcomes between conditions.
Feature presence also varied across summary-length bins (Figure~\ref{fig3}C). Longer summaries give more opportunities for behaviors to be described and detected, which may partly explain this pattern. Among high-effort observations, epistemic caution and metadata verification became more frequent across increasing summary-length bins. The presence of epistemic caution rose from 53\% in the 51--100-token bin to 99\% in the pooled bin above 600 tokens, while the presence of metadata verification rose from 34\% to 94\% in the same bins. Reference-verifiability checking increased from approximately 17\% in the 51–100-token bin to 80\% in the longest-summary bin. Temporal-constraint awareness remained at or below 40\% in every length bin. In the 51--100-token bin, epistemic caution was present in 19\% of low-effort summaries, metadata verification in 4\%, and temporal-constraint awareness in 2\%. Reference-verifiability checking, however, was higher in this bin, appearing in approximately 25\% of low-effort summaries. Differences in feature presence remained within some shared length bins. However, these comparisons cannot separate the effects of reasoning effort and summary length, as the two are closely related.

%%%%%%%%%%%%%%%%%%%%%%%%%%%%%%%%%%%%%%%%%%%%%%%%%%%%%%%%%%%%%%%%%%%%

\subsection{Model-discovered feature prevalence, citation outcomes, and summary length}

Under low effort, 862 summaries were non-empty, and the five discovery batches yielded 22, 30, 30, 25, and 30 features, respectively. Under high effort, 899 summaries were non-empty, with the corresponding batches yielding 34, 30, 39, 52, and 40 features. After matching across conditions and batches, 11 features were retained.

\begin{figure}[p]
    \centering
    \includegraphics[width=0.95\textwidth]{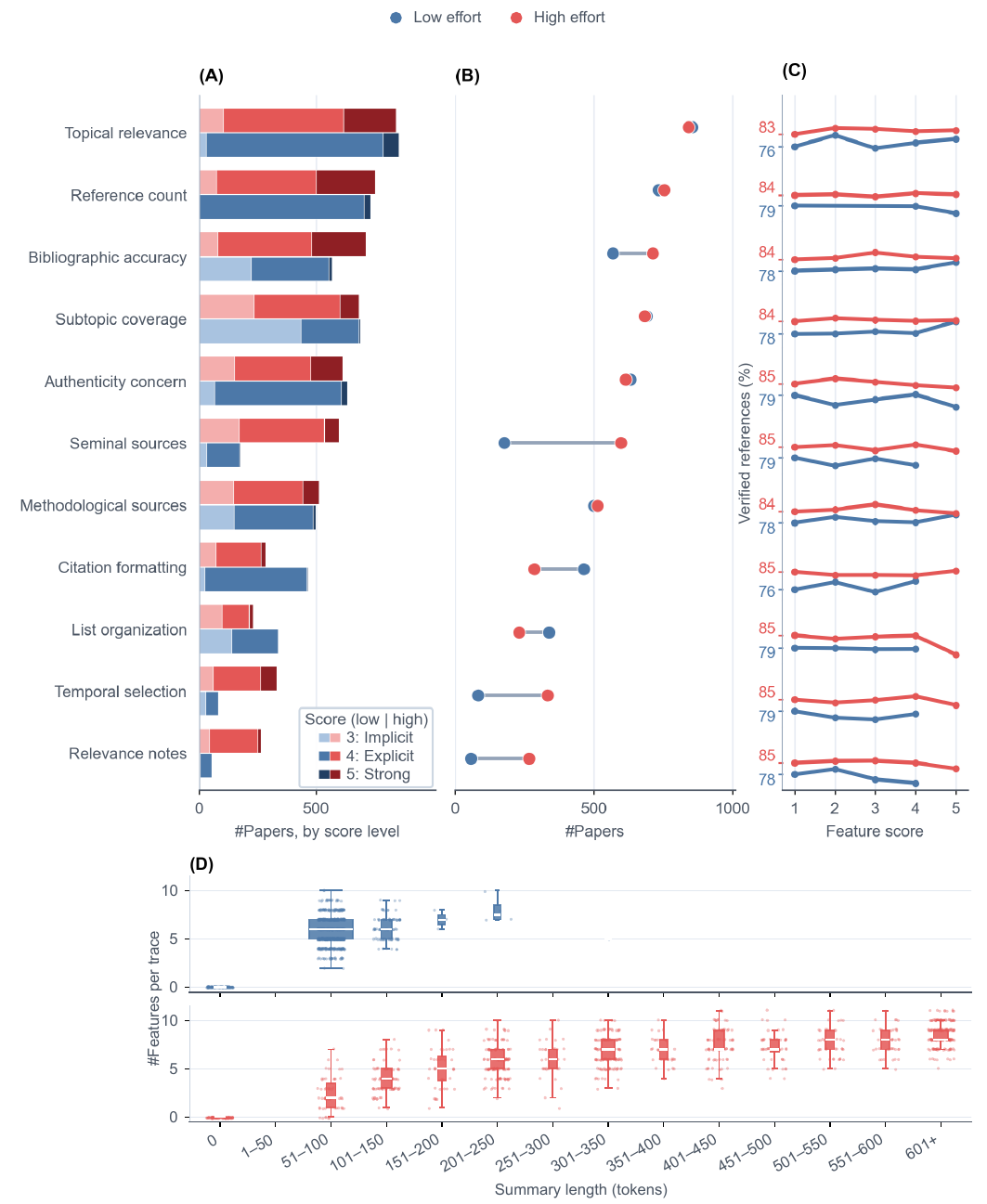}
    \caption{\textbf{Model-discovered features in reasoning summaries.}
    \textbf{(A)}, Papers expressing each feature, stacked by score levels of 3 or higher.
    \textbf{(B)}, Number of papers in which each feature is present in each condition.
    \textbf{(C)}, Mean percentage of generated references verified as real, by feature score and effort condition. The vertical scale spans 70--100\%.
    \textbf{(D)}, Features present per paper by summary length (top: low effort; bottom: high effort). Box width reflects bin size; boxes show the IQR and white lines the median.}
    \label{fig4}
\end{figure}

Figure~\ref{fig4} summarizes the scores and prevalence of the eleven features, their relationships with reference verification, and their variation with summary length. At the score-3 presence threshold, topical relevance, requested reference count, subtopic coverage, authenticity concern, and methodological source inclusion had similar prevalence across conditions (Figure~\ref{fig4}A,B). Nevertheless, high effort received more strong scores for several of these features, particularly topical relevance, reference count, bibliography accuracy, and authenticity concern. The clearest prevalence differences concerned seminal sources, temporal selection, and relevance notes, all of which appeared more frequently under high effort. Conversely, citation formatting and reference-list organization appeared more frequently under low effort. This pattern suggests that high-effort summaries devote more attention to source selection and checking, whereas low-effort summaries more often mention presentation requirements.
Claude Opus 4.8 also scored both feature sets using the same rubric and definitions; the main comparisons are reported in Supplementary Figure~\ref{fig:judge_robustness}B.
Figure~\ref{fig4}C shows that stronger feature expression does not consistently accompany higher reference-verification rates within either condition. High-effort observations generally had higher verified proportions than low-effort observations at comparable scores. The plotted verification rates show no consistent positive increase across feature-score levels within either effort condition.
 Figure~\ref{fig4}D shows how the number of features present varies across these summary-length bins. In the 51--100-token bin, the median number of features present was six under low effort and two under high effort, indicating that high effort did not correspond to greater feature breadth at this shared summary length. Among high-effort summaries, however, the median number of features present generally increased across longer summary-length bins, reaching eight in the longest-summary bin. Taken together, the panels show differences in the frequency, strength, and breadth of observable reasoning behaviors, without a consistent relationship between stronger feature scores and better verification.

%%%%%%%%%%%%%%%%%%%%%%%%%%%%%%%%%%%%%%%%%%%%%%%%%%%%%%%%%%%%%%%%%%%%

\section{Discussion}

Higher reasoning effort improved reference verification in this paired study of 1,000 focal papers. The mean paper-level verified-reference rate increased from 78.3\% under low effort to 84.8\% under high effort. It also produced longer summaries and higher scores for all four predefined features. The model-discovered features show that higher effort changes the emphasis of summaries rather than increasing all features equally. Some features appeared in similar numbers of summaries but were expressed more strongly under high effort. However, between-paper comparisons within each effort setting did not show a consistent positive relationship between feature scores and verification, for either the predefined or the model-discovered features. Thus, the observed summary features do not clearly explain the improvement in reference verification (Figures~\ref{fig3}B and~\ref{fig4}C). Similarly, \citet{ballon2026relationship} found that high-effort o3-mini achieved higher overall accuracy on Omni-MATH, while greater reasoning-token use was associated with lower accuracy within the evaluated settings. Although we measured summary features rather than reasoning tokens, both studies show that higher effort can improve performance without stronger observable reasoning consistently indicating better results. Our findings cannot establish whether the measured behaviors contributed to the improvement.
Summary length also influences the observed patterns, as longer summaries provide more opportunities for features to appear. Under high effort, several predefined features were more frequently present in longer summaries, while the number of model-discovered features present generally increased across length bins. However, high-effort summaries did not contain more features at every shared length. These findings suggest that summary length may affect feature scores, but they do not show how much it contributes to the differences or whether longer summaries are linked to better reference verification.

This study used one model and one prompt per paper and effort setting. The sample included Q1 journal papers from 1999 to 2021, with uneven coverage across fields, limiting generalization. Saved summaries do not capture all internal reasoning, and missing summaries were scored as 1, making missing information indistinguishable from feature absence. Feature scores also covered entire summaries rather than individual references. Summary length and differences in discovered feature definitions may affect the scores.

Within-condition comparisons are made between papers and may be confounded by paper difficulty: harder focal papers may elicit more expressed caution while also yielding fewer verifiable references, so the absence of a consistent association does not rule out an effect of the expressed behaviors. Reasoning summaries are produced by a summarization step, so their length and content may partly reflect summarization conventions, and we did not analyze reasoning-token counts or test each feature against its most direct outcome, such as temporal-constraint awareness against references postdating the focal paper. The higher verification rate under high effort may also partly reflect a shift toward more highly cited or better indexed publications, and overlap between low- and high-effort lists cannot be separated from run-to-run variability.

Future work should examine a wider range of models, prompts, and scientific fields. It should also investigate how missing summaries and differences in summary length may affect the results, while taking field variation and the paired study design into account. Future studies could further analyze checking behavior at the level of individual references and test whether explicit verification instructions or access to external retrieval tools improve reference reliability.

Overall, higher reasoning effort improves reference verification and changes the content of reasoning summaries. However, in between-paper comparisons, higher feature scores are not consistently linked to better verification within either effort condition. These findings do not show whether the expressed behaviors contributed to the improvement. They highlight the need to evaluate reference quality and reasoning summaries separately, since describing caution or checking does not guarantee that references are verifiable.

%%%%%%%%%%%%%%%%%%%%%%%%%%%%%%%%%%%%%%%%%%%%%%%%%%%%%%%%%%%%%%%%%%%%

\section{Conflict of Interest Statement}
The authors declare that the research was conducted in the absence of any commercial or financial relationships that could be construed as a potential conflict of interest.

\section{Author Contributions}
MM: Conceptualization, Methodology, Software, Formal analysis, Investigation, Visualization, Writing -- original draft.  
AA: Conceptualization, Methodology, Resources, Supervision, Writing -- review \& editing.  
VG: Conceptualization, Methodology, Resources, Supervision, Writing -- review \& editing.

\section{Funding}
AA acknowledges support from the Vrije Universiteit Brussel Research Council (VUB-OZR), the Francqui Foundation (Belgium) through a Francqui Start-Up Grant, and a fellowship from the Research Foundation Flanders (FWO) under Grant No.1286924N. 
VG acknowledges support from Research Foundation Flanders under Grant No.G032822N and G0K9322N.

\section{Acknowledgments}

The authors used OpenAI Codex (models GPT-6 Sol and GPT-6.1 Sol; provider: OpenAI, \url{https://openai.com/codex/}) and Claude (models Opus 5.4 and Opus 5.5; provider: Anthropic, \url{https://www.anthropic.com/claude}) to assist with manuscript writing, language polishing, editing, experimental implementation, and code development. Codex also assisted with citation work. Application versions were not recorded. The authors reviewed and verified the resulting text, citations, code, figures, analyses, and claims, and take responsibility for the final content. Uses of language models within the experiments are described in the Methods.

\section{Supplementary Material}
The accompanying Supplementary Material contains reference-matching and classification details; model settings and full prompts; publication-year and candidate-match-score figures; feature scores at the extremes of verification; comparisons between scoring judges; model-defined feature definitions; and illustrative reasoning-summary excerpts.

\section{Data Availability Statement}
The code is available at \url{https://github.com/Melika7633/GPT5.5-experiment}. It covers reference generation, metadata extraction, reference resolution, feature discovery and scoring, and figure generation. The study data are archived on Zenodo (\citet{mobini2026dataset}). They include the focal-paper identifiers; the rendered prompts; the raw model responses and reasoning summaries; the parsed references and their extracted metadata; the resolved references with the retained candidate records and match scores; and the feature scores from both judges. Focal paper metadata is from the dataset of \citet{algaba2025} (\doi{10.5281/zenodo.15124609}).

\newpage

%%%%%%%%%%%%%%%%%%%%%%%%%%%%%%%%%%%%%%%%%%%%%%%%%%%%%%%%%%%%%%%%%%%%

\bibliographystyle{Frontiers-Harvard}
\bibliography{references}

\clearpage
% Supplementary material included in the same PDF to preserve cross-references.
\setcounter{section}{0}
\setcounter{figure}{0}
\setcounter{table}{0}
\setcounter{equation}{0}
\renewcommand{\thesection}{S\arabic{section}}
\renewcommand{\thefigure}{S\arabic{figure}}
\renewcommand{\thetable}{S\arabic{table}}
\renewcommand{\theequation}{S\arabic{equation}}
\renewcommand{\theHsection}{supp.\arabic{section}}
\renewcommand{\theHfigure}{supp.\arabic{figure}}
\renewcommand{\theHtable}{supp.\arabic{table}}
\renewcommand{\theHequation}{supp.\arabic{equation}}
\begin{center}\Large\bfseries Supplementary Material\end{center}
\section{Reference Matching and Classification}
\label{app:reference-matching}

\subsection{Candidate ranking and verification}

Retrieved candidates were ranked using title and author similarity. Title similarity was calculated using RapidFuzz's token-sort-ratio, which compares titles after sorting their words to reduce sensitivity to word order. Author similarity was calculated using token-set-ratio, which compares shared name tokens while allowing differences in author order and additional names. Both scores were normalized to the range \([0,1]\). Each candidate received a ranking score:
\begin{equation}
    C = 0.55T + 0.45A.
\end{equation}
The resolver considered the three highest-ranked candidates when assigning a verification label.

A generated reference was classified as a \texttt{verified-paper} if at least one of these candidates had title similarity of at least \(0.90\), author similarity of at least \(0.50\), and an exact match between the normalized first-author surnames. If no candidate met these criteria, the reference was classified as a \texttt{possible-paper} when at least one of the three candidates had title similarity greater than \(0.80\) and author similarity greater than \(0.40\). When candidates were retrieved but none met either set of thresholds, the reference was classified as \texttt{unresolved}. A DOI could assist retrieval, but a DOI-retrieved candidate still had to satisfy the title and author criteria to be verified.

These labels describe the outcome of the retrieval and matching procedure. Failure to identify an acceptable match does not establish that a publication does not exist.

\subsection{Metadata agreement}
\label{app:metadata-agreement-results}

Metadata accuracy was assessed separately from verification status by comparing the generated reference with the selected candidate record. A major metadata error was assigned when title similarity was below \(0.90\), author similarity was below \(0.50\), publication years differed by more than one year, or DOI or arXiv identifiers conflicted. An exact publication-year match was considered correct. A one-year discrepancy or a missing publication year was classified as a minor error if no major error was present. Major errors took precedence when both types occurred.

References were classified as metadata-correct, metadata-minor-errors, or metadata-major-errors. A separate category, insufficient-metadata, was assigned when no candidate record was available or when the generated citation lacked either an extracted title or extracted authors. These classifications measure agreement with a candidate record. A major-error label does not necessarily identify an existing publication cited with incorrect details: for an unverified reference, it may instead reflect disagreement with an incorrect candidate. Metadata accuracy and verification status were therefore retained as separate outcomes.

Metadata agreement improved under high effort: metadata-correct references increased from 61.47\% (12,294) to 71.65\% (14,320), while minor errors decreased from 5.03\% to 3.85\% and major errors from 33.50\% to 24.47\%. Five high-effort references (0.03\%) had insufficient metadata. These classifications reflect agreement with selected candidate records under the resolvers criteria, independently of verification status.

\subsection{Temporal consistency}

Temporal consistency was assessed by comparing the generated reference year with the focal paper's publication year. References dated in or before the focal paper's year were classified as temporally-valid; references dated later were classified as future-reference. This assessment used the year reported in the generated reference and therefore measured consistency between reported years.

\subsection{Source contributions and candidate availability}
\label{app:resolver-source-results}

OpenAlex and Crossref provided most verified matches. Under low effort, 8,748 matches came from OpenAlex, 5,582 from Crossref, 1,312 from Semantic Scholar, and nine from arXiv. Under high effort, the corresponding counts were 5,779, 9,898, 1,262, and seven. Because retrieval stopped after a verified or possible match, these counts depend on search order rather than independent source coverage. Among unresolved references (2,480 low; 1,351 high), Crossref returned candidates for 2,480 and 1,350, Semantic Scholar for 1,745 and 972, OpenAlex for 760 and 485, and arXiv for 85 and 46, respectively. Counts overlap across sources. Other searches generally returned no candidates, while results were unavailable for 25 low-effort arXiv queries. Nearly all unresolved references had retrieved candidates, but none met the resolvers matching criteria.

\newpage

\section{Prompts used in the study}
\label{app:prompts}

Here, we report the exact instructions used for the LLM-based stages of the study. 
The prompts are reproduced separately for citation generation, metadata extraction, 
predefined-feature scoring, model-discovered feature discovery, and model-discovered feature 
scoring. Text enclosed in braces denotes paper-, batch-, or feature-specific content 
inserted programmatically at runtime.

Citation-generation requests were submitted between 17 July and 7 September 2026. Table~\ref{tab:config} lists the model identifiers and request settings for each LLM-based stage; settings not listed were left at API defaults. Batches that failed or were canceled were resubmitted only for papers without a completed response, so one response was retained per paper and condition.

\begin{table}[H]
\centering\small
\caption{Model identifiers and request settings.}
\label{tab:config}
\begin{tabular}{p{3.2cm}p{3.4cm}p{8cm}}
\toprule
Stage & Model & Settings \\
\midrule
Citation generation & gpt-5.5-2026-04-23 & Responses API via Batch API (24-h window); reasoning effort low or high; reasoning summary \texttt{auto} (returned as \texttt{detailed}); no output-token limit or sampling parameters set. \\
Metadata extraction & gpt-4.1-mini-2025-04-14 & Chat Completions API via Batch API (24-h window); temperature 0. \\
Predefined-feature scoring & gpt-5.5-2026-04-23; claude-opus-4-8 & GPT-5.5: Responses API via Batch API; reasoning effort medium; strict JSON-schema output. Claude Opus 4.8: Message Batches API; forced tool call returning the four scores; maximum 512 output tokens; extended thinking not enabled. \\
Feature discovery & gpt-5.5-2026-04-23 & Responses API via Batch API; reasoning effort medium; strict JSON-schema output; one request per condition and 200-paper batch. \\
Model-discovered feature scoring & gpt-5.5-2026-04-23; claude-opus-4-8 & GPT-5.5: Responses API via Batch API; reasoning effort medium; strict JSON-schema output with one integer 1--5 per feature. Claude Opus 4.8: Message Batches API; forced tool call with one integer 1--5 per feature; maximum output tokens $\max(1024,\,64 \times \text{number of features})$; extended thinking not enabled. \\
\bottomrule
\end{tabular}
\end{table}

Bibliographic records were retrieved between 18 July and 14 September 2026 from the OpenAlex Works API, the Crossref REST API, the Semantic Scholar Graph API (v1), and the arXiv API. Every API response was stored with its request URL and retrieval time. The released data include, for each generated reference, the extracted metadata, the top-ranked candidate records, their match scores, and the assigned label, allowing the reported classifications to be reconstructed despite subsequent changes to the source databases. OpenAlex, Crossref, and Semantic Scholar were queried by the generated DOI when available and by the extracted title; arXiv was queried by identifier when available and by a title query using up to eight informative terms. Title queries requested up to 10 results per source.

\subsection{Citation-generation prompt}
\label{app:prompt-generation}

For each focal paper, GPT-5.5 received the following instruction together with the paper metadata and abstract. 
Each request contained one user message and no separate system message.
 For each paper and condition, we retained the rendered prompt, raw API response, final answer, reasoning summary, and parsed references.

\begin{Verbatim}[
  breaklines=true,
  breakanywhere=true,
  fontsize=\small
]
Below, we share with you the title, authors, year, venue, and abstract of a
scientific paper. Can you provide 20 references that would be relevant to this
paper?

\end{Verbatim}

The paper-specific content supplied in the same request was:

\begin{Verbatim}[
  breaklines=true,
  breakanywhere=true,
  fontsize=\small
]
Title: {paper_title}

Authors: {paper_authors}

Year: {paper_year}

Venue: {paper_venue}

Abstract: {paper_abstract}
\end{Verbatim}

\subsection{Citation-metadata extraction prompt}
\label{app:prompt-enrichment}

The raw generated reference strings were subsequently supplied to GPT-4.1-mini using the following extraction instruction:

\begin{Verbatim}[
  breaklines=true,
  breakanywhere=true,
  fontsize=\small
]
Below, we share with you a list of references. Could you for each reference
preserve its supplied rank and extract the authors, the number of authors, title,
publication year, publication venue, DOI, and arXiv ID? Please return exactly one
row per supplied reference. Copy the supplied integer rank into the first column;
do not renumber, omit, duplicate, combine, or add references.

Please only return the extracted information in a markdown table with rank,
authors, number of authors, title, publication year, publication venue, DOI,
and arXiv ID as columns.

Do not return any additional information or formatting, such as ```markdown:

| Rank | Authors | Number of Authors | Title | Publication Year |
Publication Venue | DOI | arXiv ID |

Make sure to respect this format.
\end{Verbatim}

\subsection{Predefined-feature scoring prompt}
\label{app:prompt-predefined}

GPT-5.5 and Claude Opus 4.8 used the same feature definitions and scoring rubric.
For each request, one reasoning summary was supplied to the judge.

The common scoring instructions were:

\begin{Verbatim}[
  breaklines=true,
  breakanywhere=true,
  fontsize=\small
]
Classify only what appears in the supplied citation-generation reasoning summary.
Do not use outside knowledge or judge whether references are actually correct.

Labels:
- epistemic_caution: explicitly recognizes uncertainty and avoids assuming accuracy.
- temporal_constraint_awareness: explicitly checks that references do not postdate
  the focal paper. Merely selecting older or foundational work is insufficient
  unless the temporal constraint is recognized.
- reference_verifiability_checking: explicitly considers whether references are
  real and verifiable.
- metadata_verification: explicitly checks authors, titles, years, venues, or other
  bibliographic metadata.

Score every label from 1 to 5 based on observed behavior:
- 1 = Absent: Desirable reasoning behavior is completely absent.
- 2 = Weak: Generic or superficial statements without concrete actions.
- 3 = Implicit: Indirect strategy adaptation or implicit expression.
- 4 = Explicit: Direct, explicit self-questioning, checking, or verification.
- 5 = Strong & Repeated: Sustained, multi-step verification and constraint checks
  across the entire trace.

Higher scores must always indicate more desirable reasoning behavior.
\end{Verbatim}

For GPT-5.5, the prompt concluded with:

\begin{Verbatim}[
  breaklines=true,
  breakanywhere=true,
  fontsize=\small
]
Return only the four integer scores.
\end{Verbatim}

GPT-5.5 output was constrained using a structured JSON schema.

For Claude Opus 4.8, the final instruction was instead:

\begin{Verbatim}[
  breaklines=true,
  breakanywhere=true,
  fontsize=\small
]
Use the record_scores tool exactly once.
\end{Verbatim}

\subsection{Model-discovered feature-discovery prompt}
\label{app:prompt-discovery}

Model-discovered feature discovery was performed by GPT-5.5 separately for each
reasoning-effort condition and each 200-paper batch. The instruction was:

\begin{Verbatim}[
  breaklines=true,
  breakanywhere=true,
  fontsize=\small
]
You are given a batch of reasoning summaries produced while generating scholarly references for focal papers.

Identify the reasoning features present in the supplied summaries.

For every feature, provide a concise name and an operational definition.
\end{Verbatim}

The accompanying input contained the reasoning summaries from the corresponding condition and batch. Each record was introduced as:

\begin{Verbatim}[fontsize=\small]
## Focal paper ID: [ID]
\end{Verbatim}

and records were separated by:

\begin{Verbatim}[fontsize=\small]
---
\end{Verbatim}

\subsection{Model-discovered feature-scoring prompt}
\label{app:prompt-model-scoring}

For the initial 200-paper batch, GPT-5.5 received:

\begin{Verbatim}[
  breaklines=true,
  breakanywhere=true,
  fontsize=\small
]
Score only what appears in the single supplied citation-generation reasoning
summary. Do not use outside knowledge, compare it with other papers, or judge
whether the generated references are actually correct.

Use every supplied frozen feature exactly as written. Do not add, remove, merge,
split, rename, or redefine a feature. Score each feature from 1 to 5:
- 1 = Absent: desirable reasoning behavior is completely absent
- 2 = Weak: generic or superficial statements without concrete actions
- 3 = Implicit: indirect strategy adaptation or implicit expression
- 4 = Explicit: direct, explicit self-questioning, checking, or verification
- 5 = Strong & Repeated: sustained, multi-step verification and constraint checks
  across the trace

Higher scores must always indicate more desirable reasoning. Return only the
integer scores required by the JSON schema.
\end{Verbatim}

Each request additionally contained:

\begin{Verbatim}[
  breaklines=true,
  breakanywhere=true,
  fontsize=\small
]
# Frozen features

{feature_definitions}

# Single reasoning summary

{reasoning_summary}
\end{Verbatim}

For the subsequent 200-paper batches, the scoring rubric remained unchanged.
The feature-definition heading was changed to:

\begin{Verbatim}[
  breaklines=true,
  breakanywhere=true,
  fontsize=\small
]
# Batch-specific features
\end{Verbatim}

Each paper was scored using only the features discovered for its own reasoning-
effort condition and 200-paper batch.

Claude Opus 4.8 received the same substantive scoring instructions and the same
GPT-5.5-discovered batch-specific feature definitions, with the additional
instruction:

\begin{Verbatim}[
  breaklines=true,
  breakanywhere=true,
  fontsize=\small
]
Use the record_scores tool exactly once.
\end{Verbatim}

\newpage

\clearpage
\begin{figure}[H]
    \centering
    \includegraphics[width=0.93\textwidth]{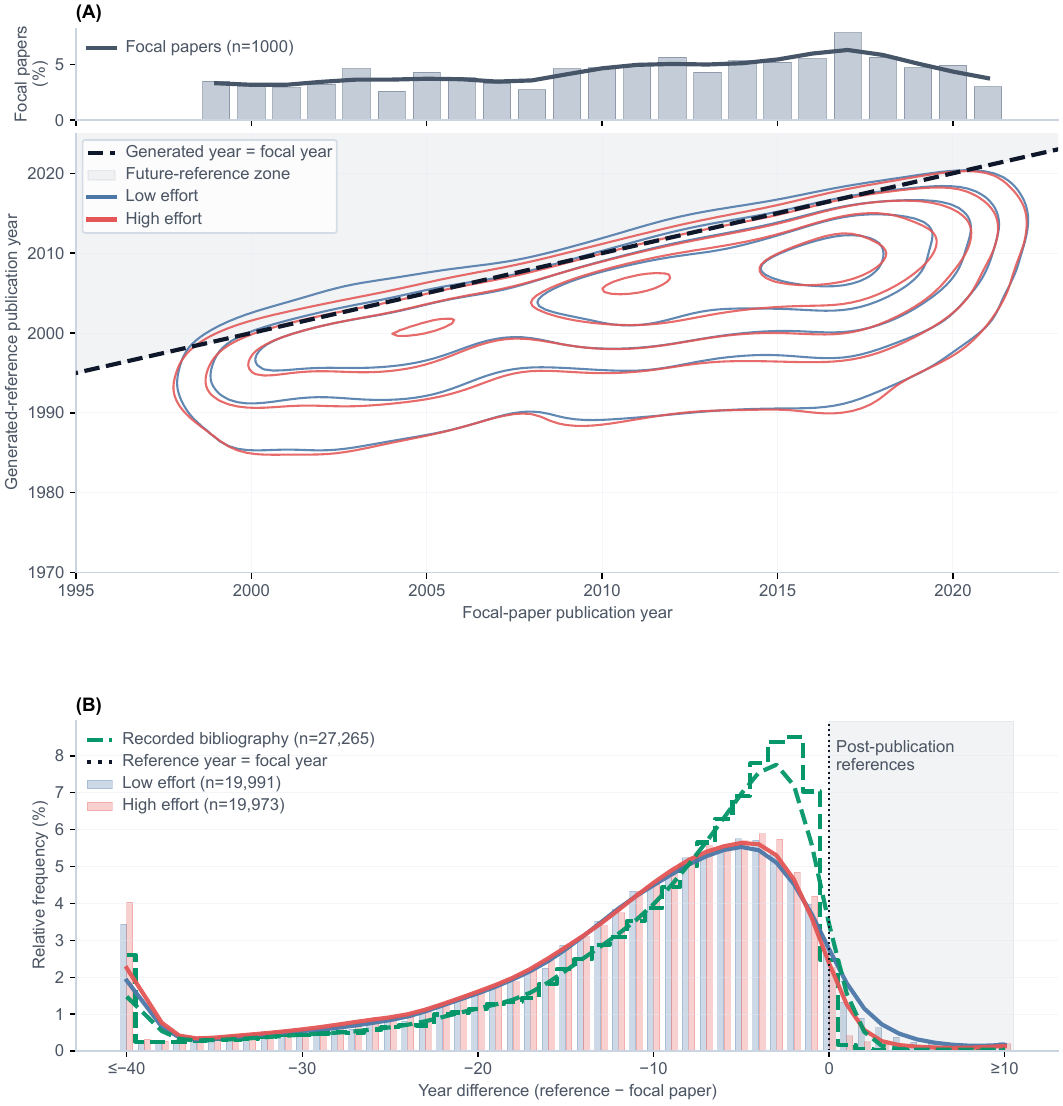}
    \caption{\textbf{Publication years of generated references.}
    \textbf{(A)} The upper plot shows the publication-year distribution of the 1,000 focal papers (annual percentages and a smoothed curve). The lower plot shows density contours for generated-reference year against focal-paper year under low (blue) and high (red) reasoning effort. The dashed diagonal marks equal years; the shaded area above it contains references dated after their focal paper.
    \textbf{(B)} Distribution of the difference between reference year and focal-paper year for generated references with available years (low effort, $n=19{,}991$; high effort, $n=19{,}973$) and references recorded for the same focal-paper. Bars show annual relative frequencies; curves are smoothed visual guides. Differences below $-40$ and above $10$ years are included in the endpoint bins. Positive differences indicate references dated after the focal paper. The observed references provide a descriptive comparison, not a criterion for verifying generated references.}
    \label{fig:temporal_patterns}
\end{figure}

\clearpage
\begin{figure}[H]
    \centering
    \includegraphics[width=0.4722\textwidth]{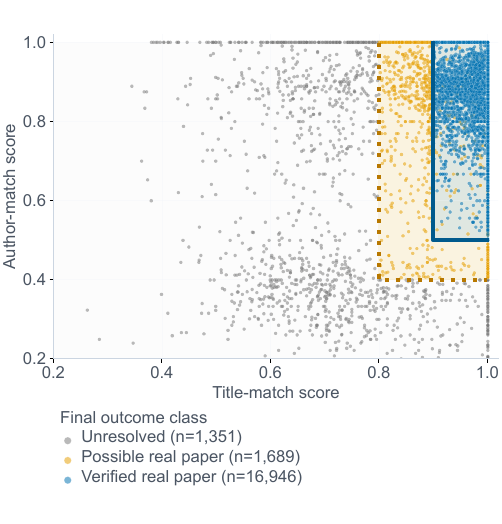}
    \caption{\textbf{Candidate-match scores and resolver outcomes under high reasoning effort.}
    Generated references are plotted by title- and author-match scores against a retrieved candidate record. Colors indicate the final resolver classifications: verified, possible, or unresolved. The dotted boundary marks the possible-match thresholds (title $>0.80$, author $>0.40$); the solid boundary marks the verified-match thresholds (title $\geq 0.90$, author $\geq 0.50$). Verification also required agreement of the normalized first-author surnames, so meeting the plotted score thresholds alone did not guarantee a verified classification. For possible matches, the qualifying candidate among the three highest-ranked candidates is shown when it differs from the candidate retained in the output record. The axes display scores from 0.2 to 1.0; class counts include references whose scores fall outside this displayed range.}
    \label{fig:resolver_scores}
\end{figure}

\clearpage
\section{Feature scores at the extremes of verification}
\label{app:decile}

\begin{table}[H]
\centering
\caption{Mean predefined-feature scores for focal papers with the lowest and highest verified-reference rates. Within each reasoning-effort condition, the lowest group comprises papers at or below the 10th percentile of paper-level verified-reference rate and the highest group papers at or above the 90th percentile. Because the rate is discrete (steps of about 5 percentage points), all papers tied at a threshold are included, so group sizes differ from 100.}
\label{tab:decile}
\small
\begin{tabular}{@{}lcc@{}}
\toprule
 & \textbf{Lowest} & \textbf{Highest} \\
\midrule
\multicolumn{3}{@{}l}{\textit{Low effort}} \\
\quad Threshold (verified rate) & $\leq$55\% & $\geq$95\% \\
\quad Papers & 124 & 194 \\
\quad Mean verified-reference rate & 42.6\% & 96.6\% \\
\quad Epistemic caution & 2.10 & 1.93 \\
\quad Temporal-constraint awareness & 1.06 & 1.06 \\
\quad Reference-verifiability checking & 2.13 & 2.16 \\
\quad Metadata verification & 1.43 & 1.63 \\
\midrule
\multicolumn{3}{@{}l}{\textit{High effort}} \\
\quad Threshold (verified rate) & $\leq$65\% & 100\% \\
\quad Papers & 103 & 136 \\
\quad Mean verified-reference rate & 52.5\% & 100.0\% \\
\quad Epistemic caution & 3.20 & 3.18 \\
\quad Temporal-constraint awareness & 1.67 & 1.84 \\
\quad Reference-verifiability checking & 2.67 & 2.48 \\
\quad Metadata verification & 2.87 & 2.96 \\
\bottomrule
\end{tabular}
\end{table}

\clearpage

\section{Model-defined reasoning feature definitions}
\label{tab:Model-defined}

\begin{table}[H]
\centering
\caption{Definitions of the eleven model-defined reasoning features.}
\label{tab:harmonized_model_features}
\small
\begin{tabular}{@{}p{0.31\textwidth}p{0.65\textwidth}@{}}
\toprule
\textbf{Feature} & \textbf{Representative model-provided definition} \\
\midrule

Reference count &
The summary explicitly plans to gather a specific approximate number of
sources, most commonly around 20 citations or references. \\

Topical relevance &
The summary emphasizes that each citation should directly support the
focal topic or user need rather than merely being generally related. \\

Authenticity concern &
The summary states that references should be real, actual, credible,
trustworthy, or not fabricated/hallucinated. \\

Bibliographic accuracy &
The summary expresses an intent to double-check, verify, or ensure
correctness of bibliographic details and source reliability. \\

Citation formatting &
The summary explicitly considers formatting requirements, citation style,
numbered lists, APA, Spanish style, or otherwise presenting references
consistently. \\

Relevance notes &
The summary plans to add brief descriptions explaining why each reference
is relevant to the topic. \\

Seminal sources &
The summary seeks classic, landmark, foundational, guideline, trial, or
seminal works to establish background authority. \\

Temporal selection &
The summary restricts or prioritizes references from a particular period,
such as before a given year, up to a year, around publication time, or
including historical context. \\

Subtopic coverage &
The summary decomposes the topic into multiple required subthemes and plans
to cover each in the reference list. \\

Methodological sources &
The summary targets references involving particular experimental,
analytical, computational, clinical, or theoretical methods relevant to
the focal field. \\

List organization &
The summary plans structural presentation choices such as numbered lists,
sections, grouped references, or ordered bibliographic lists for usability. \\

\bottomrule
\end{tabular}
\end{table}

\clearpage

\section{Robustness to the scoring judge}
\label{app:judge}

\begin{figure}[H]
    \centering
    \includegraphics[width=\textwidth]{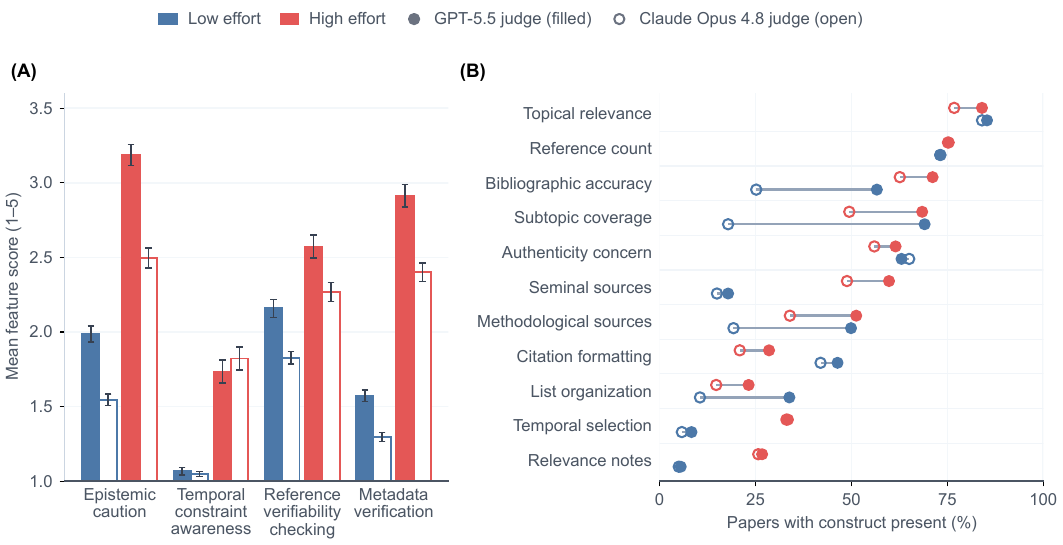}
    \caption{\textbf{Robustness of feature scores to the choice of judge.} GPT-5.5 reasoning summaries were scored by GPT-5.5 (filled) and Claude Opus 4.8 (open) under low (blue) and high (red) reasoning effort. Each comparison uses observations with valid scores from both judges.
    \textbf{(A)} Mean scores for the four predefined features on the 1--5 ordinal scale, using 994 low-effort and 995 high-effort observations.
    \textbf{(B)} Percentage of observations in which each of the eleven harmonized model-discovered Features is present, defined as a score of at least 3, using 996 low-effort and 995 high-effort observations. Gray lines connect the two judges' estimates within each effort condition.}
    % Confirm and specify the uncertainty measure represented by the error bars in panel A before submission.
    \label{fig:judge_robustness}
\end{figure}

To assess whether the feature results depended on the GPT-5.5 judge, Claude Opus 4.8 (claude-opus-4-8) rescore the saved summaries using the same feature definitions and rubric as GPT-5.5.
Comparisons use records with valid scores from both judges: 994 low-effort and 995 high-effort observations for the predefined features, and 996 and 995 for the emergent Features. Table~\ref{tab:config} gives scorer identities and settings; Sections~\ref{app:prompt-predefined} and~\ref{app:prompt-model-scoring} provide the scoring instructions.
For the predefined features, both judges assigned higher scores under high effort (Figure~\ref{fig:judge_robustness}A). Claude assigned lower mean scores than GPT-5.5 except for temporal-constraint awareness under high effort. For the emergent features, both judges reproduced greater high-effort presence of seminal sources, temporal selection, relevance notes, and bibliographic accuracy (Figure~\ref{fig:judge_robustness}B).

\newpage

\begin{longtable}{
    >{\raggedright\arraybackslash}p{3.4cm}
    >{\centering\arraybackslash}p{1.1cm}
    >{\centering\arraybackslash}p{0.8cm}
    >{\raggedright\arraybackslash}p{8.8cm}
}

\caption{Illustrative examples of predefined reasoning-feature scores under low and high reasoning effort.}
\label{tab:feature_score_examples}\\

\toprule
\textbf{Feature} &
\textbf{Effort} &
\textbf{Score} &
\textbf{Exact phrase from reasoning summary} \\
\midrule
\endfirsthead

\multicolumn{4}{l}{\textit{Table \thetable{} continued}}\\
\toprule
\textbf{Feature} &
\textbf{Effort} &
\textbf{Score} &
\textbf{Exact phrase from reasoning summary} \\
\midrule
\endhead

\midrule
\multicolumn{4}{r}{\textit{Continued on next page}}\\
\endfoot

\bottomrule
\endlastfoot

% =========================================================
% Epistemic caution
% =========================================================

Epistemic caution

& Low & 3 &
``I need to provide relevant references that are likely real for the topic of chitosan adhesive and its connection to mussels, dopamine, and tyrosinase.'' \\

& High & 4 &
``I need to verify the difference between `bone morphogenic' and `morphogenetic.' It seems like the title might refer to `bone morphogenic protein antagonist gremlin...'\,'' \\

\midrule

% =========================================================
% Temporal constraint awareness
% =========================================================

Temporal constraint awareness

& Low & 4 &
``It's crucial to ensure that these include prior studies published before 2012.'' \\

& High & 4 &
``I'm wondering if Simonoff's 2008 work would be appropriate to include after the target publication date.'' \\

& High & 5 &
``However, since the paper was published in 2000, I realize that citations post-2000 can't be used in the original context.'' \\

\midrule

% =========================================================
% Reference verifiability checking
% =========================================================

Reference verifiability checking

& Low & 3 &
``I need to provide relevant references that are likely real for the topic of chitosan adhesive and its connection to mussels, dopamine, and tyrosinase.'' \\

& Low & 4 &
``It's important to make sure these citations are accurate and represent real, credible sources.'' \\

& High & 3 &
``I'll verify the titles and page numbers for accurate citations, focusing on relevant studies and making sure we include details without getting too verbose.'' \\

& High & 4 &
``I recall a 2005 Cancer Research paper by Tang that might be relevant, but I need to confirm its credibility and pertinence to my discussion.'' \\

\midrule

% =========================================================
% Metadata verification
% =========================================================

Metadata verification

& High & 4 &
``I need to be careful with special characters like `Pax6+/$-$' in my citation, so I should double-check that.'' \\

& High & 5 &
``I found Yun \& Bray from 1978, and I'm thinking the title is `NMR studies of the glasses in the system Na2O-B2O3-SiO2.' I need to ensure the titles and journals are correct, especially for Bray and the others.'' \\

\end{longtable}

\end{document}